\def\year{2019}\relax
\documentclass[letterpaper]{article} 
\usepackage{aaai19}  
\usepackage{times}  
\usepackage{helvet}  
\usepackage{courier}  
\usepackage{url}  
\usepackage{graphicx}  
\usepackage{amsmath}
\usepackage{algorithm}
\usepackage{xcolor}
\usepackage{algorithmic}
\usepackage{multirow}
\usepackage{booktabs}
\usepackage{multirow}
\usepackage{placeins}
\usepackage{tabularx}
\usepackage{threeparttable}
\usepackage{subcaption}
\usepackage{amsfonts}

\makeatletter

\newcommand{\Rmnum}[1]{\expandafter\@slowromancap\romannumeral #1@}
\makeatother

\begin{document}
%


\setlength\titlebox{2.5in}

\title{Exploiting Local Feature Patterns for Unsupervised Domain Adaptation}
\author{Jun Wen\textsuperscript{1,2}, Risheng Liu\textsuperscript{3}, Nenggan Zheng\textsuperscript{1}\thanks{Coresponding author.}, Qian Zheng\textsuperscript{2}, Zhefeng Gong\textsuperscript{4}, Junsong Yuan\textsuperscript{5}  \\
\textsuperscript{1} Qiushi Academy for Advanced Studies, Zhejiang University, Hangzhou, Zhejiang, China \\
\textsuperscript{2} College of Computer Science and Techology, Zhejiang University, Hangzhou, Zhejiang, China \\
\textsuperscript{3} International School of Information Science \& Engineering, Dalian University of Technology, Liaoning, China \\
\textsuperscript{4} Department of Neurobiology, Zhejiang University School of Medicine, Hangzhou, Zhejiang, China \\
\textsuperscript{5} State University of New York at Buffalo \\
 \{junwen, qianzheng, zfgong\}@zju.edu.cn, zng@cs.zju.edu.cn, rsliu@dlut.edu.cn, jsyuan@buffalo.edu}

\maketitle

\begin{abstract}
Unsupervised domain adaptation methods aim to alleviate performance degradation caused by domain-shift by learning domain-invariant representations. Existing deep domain adaptation methods focus on holistic feature alignment by matching source and target holistic feature distributions, without considering local features and their multi-mode statistics. We show that the learned local feature patterns are more generic and transferable and a further local feature distribution matching enables fine-grained feature alignment. In this paper, we present a method for learning domain-invariant local feature patterns and jointly aligning holistic and local feature statistics. Comparisons to the state-of-the-art unsupervised domain adaptation methods on two popular benchmark datasets demonstrate the superiority of our approach and its effectiveness on alleviating negative transfer.
\end{abstract}

\section{Introduction}

Many machine learning algorithms assume that the training and testing data are drawn from the same feature space with the same distribution. However, this assumption rarely holds in practice as the data distribution is likely to change over time and space. Though the state-of-the-art deep convolutional features show invariant to low-level variations to some degree, they are still susceptible to domain-shift, as we can not manually label sufficient training data that cover diverse application domains \cite{csurka2017domain,zhou2018transfer,hong2018conditional}. The typical solution is to further finetune the learned deep models on task-specific datasets. However, it is often prohibitively difficult and expensive to obtain enough labeled data to properly finetune the large-scale parameters employed by deep networks. Instead of recollecting labeled data and retraining the model for every possible new scenario, unsupervised domain adaptation methods attempt to alleviate the performance degradation by transferring discriminative features from neighboring labeled source domains using only unlabeled target data \cite{ganin2016domain,tzeng2017adversarial}.

\begin{figure}[!hbtp]
    \centering
        \includegraphics[width=0.39 \textwidth]{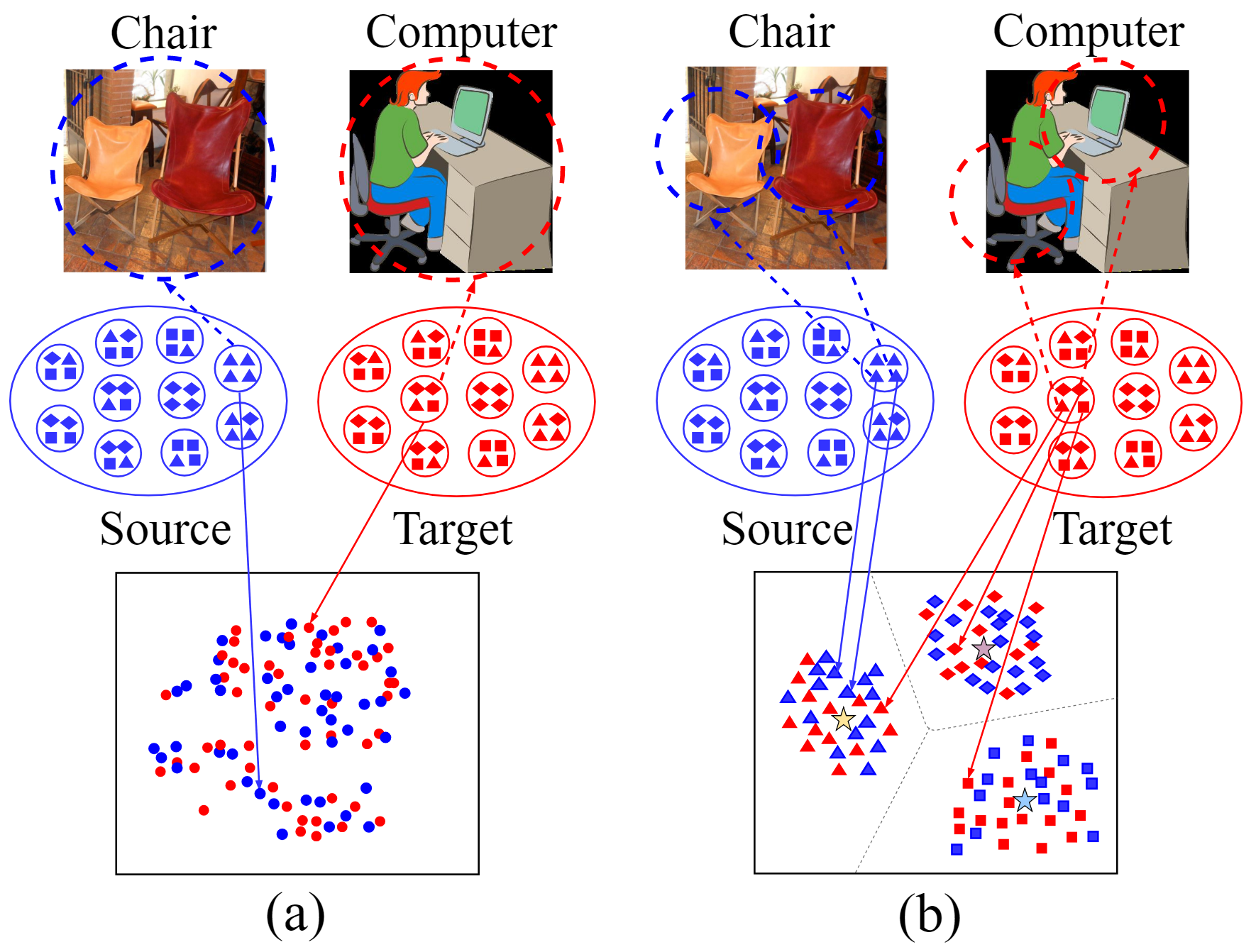}
   \caption{ Comparisons of (a) holistic feature alignment and (b) local feature alignment. The learned local feature pattern ``Chair" could not only be shared across domains but also between the category ``Chair" and category ``Computer", as shown in (b).}
\end{figure}

Unsupervised domain adaptation methods assume shared label space with different feature distributions across source and target domains. These methods usually bridge different domains by learning domain-invariant discriminative representations, and directly apply the classifier learned from only source labels to target domain \cite{ben2010theory}. To reduce domain discrepancy, previous methods usually align source and target in a shared subspace \cite{gong2012geodesic,fernando2013unsupervised}. Recently, deep neural networks have been exploited to map both domains into a domain-invariant feature space and learn more transferable representations \cite{zhou2014hybrid,tzeng2017adversarial,pei2018multi}. This is generally achieved by optimizing the learned representations to minimize some measures of domain discrepancy such as maximum mean discrepancy \cite{long2015learning}, reconstruction loss \cite{ghifary2016deep}, correlation distance \cite{sun2016deep}, or adversarial loss \cite{tzeng2017adversarial}. Among them, the adversarial learning based deep domain adaptation methods have become increasingly prevalent and achieved the top performances.

Though existing deep domain adaptation methods have achieved excellent performances \cite{csurka2017domain}, they mainly focus on aligning source and target holistic representations \cite{ganin2016domain,tzeng2017adversarial}, without exploiting the more primitive and transferable local feature patterns. Comparing to local feature patterns, holistic representations, usually the final fully-connected layers of deep neural networks, are tailored to capture more current task related semantics and hence less transferable to novel domains \cite{yosinski2014transferable}, especially when source labels are scarce. In contrast, local feature patterns only focus on smaller parts of images that could be shared not only across different domains but also between multiple categories, as shown in Figure 1(b), thus are more generic and less susceptible to limited training labels. Further, existing domain adaptation methods fail to consider the complex multi-mode distributions of local features, which limits their capability to achieve fine-grained local feature alignment.

Motivated by the above limitations of existing domain adaptation methods, we propose to learn transferable local feature patterns for unsupervised domain adaptation and jointly align holistic features and local features for fine-grained alignment. The local feature space is firstly partitioned into several well-separated cells with a cluster of generic local feature patterns. We then achieve feature alignment by simultaneously enforcing both holistic distribution consistency over the aggregated local features and local feature alignment within each separated local feature pattern cells, as shown in Figure1(b). The contributions of our work are as follows:

\begin{itemize}

\item Different from most existing domain adaptation methods which focus on aligning holistic features, we propose to exploit local features for unsupervised domain adaptation. We show that our learned local feature patterns are more generic and transferable.

\item We align the residuals of local features regarding to the learned local feature patterns by minimizing an additional conditional domain adversarial loss. With joint holistic and local distribution matching, we enable fine-grained cross-domain feature alignment.

\item  Exhaustive experimental results on standard domain adaptation benchmarks demonstrate the promises of the proposed method by outperforming the state-of-the-art approaches. As a nontrivial byproduct, we provide comprehensive evaluations of local feature patterns from different levels for unsupervised domain adaptation.

\end{itemize}

\section{Related Works}

We first give a brief overview on existing domain adaptation methods. Then, we present related works on local feature aggregation.

\noindent \textbf{Domain Adaptation} methods seek to learn from neighbouring source domains discriminative representations that can be applied to target domains. This is usually achieved by mapping samples from both domains into a domain-invariant feature space to reduce domain discrepancy \cite{ben2010theory}. Previous methods usually seek to align source and target feature through subspace learning \cite{gong2012geodesic,fernando2013unsupervised,pan2011domain}. Recently deep domain adaptation approaches become prevalent as deep networks can learn more transferable representations \cite{bengio2013representation,yosinski2014transferable}. Different measures of domain discrepancy have been minimized to align source and target distributions. Several methods propose to minimize the Maximum Mean Discrepancy (MMD) loss between source and target \cite{long2015learning}. Ghifary et al. propose to reduce the discrepancy through the auto-encoder based reconstruction loss \cite{ghifary2016deep}. Recently, the adversarial learning based methods are becoming popular \cite{ganin2016domain,tzeng2017adversarial,pei2018multi,zheng2018unsupervised}. These methods are closely related to the adversarial generative networks (GAN) \cite{goodfellow2014generative,gulrajani2017improved}. They aim to reduce domain discrepancy by optimizing the feature learning network with an adversarial objective produced by another discriminator network which is trained to distinguish features of target from features of source. All these methods only focus on transferring holistic semantics, without considering the more generic local feature patterns and the multi-mode statistics of local features.

\noindent \textbf{Feature Aggregation} Our work is also related to feature aggregation methods, such as vectors of locally aggregated descriptors (VLAD) \cite{jegou2010aggregating}, bag of visual words (BoW) \cite{sivic2003video}, and Fisher vectors (FV) \cite{perronnin2007fisher}. Previously, these methods have usually been applied to aggregate hand-crafted keypoint descriptors, such as SIFT, as a post-processing step, and only recently have them been extended to encode deep convolutional features with end-to-end training \cite{arandjelovic2016netvlad}. VLAD has been successfully applied to image retrieval \cite{yue2015exploiting}, place recognition \cite{arandjelovic2016netvlad}, action recognition \cite{girdhar2017actionvlad}, etc. We build on the end-to-end trainable VLAD, and extend it to learn generic local feature patterns and facilitate local feature alignment for unsupervised domain adaptation.

\begin{figure*}[!hbtp]
    \centering
        \includegraphics[width=0.85 \textwidth]{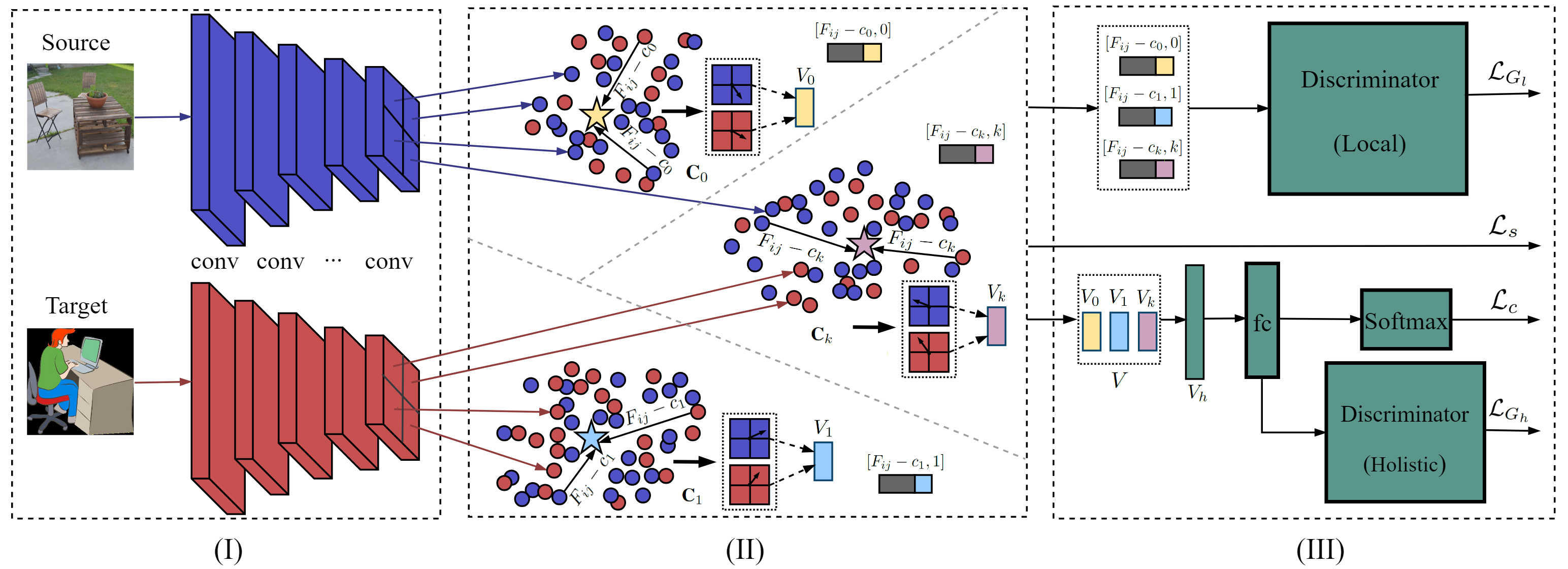}
   \caption{Pipeline of the proposed method: (\Rmnum{1}) feature extractor $\textbf{G}$, (\Rmnum{2}) local feature patterns learning, and (\Rmnum{3}) feature alignment. We learn generic local feature patterns, and jointly align holistic features and local features.}
\end{figure*}

\section{Method}
In this section, we describe the proposed unsupervised domain adaptation method. Given source domain dataset $\mathcal{D}_s=\left\{(x_{i}^{s},y_{i}^{s})\right\}_{i=1}^{n_s}$ of $n_s$ labeled examples and target domain dataset $\mathcal{D}_t=\left\{x_{j}^{t}\right\}_{j=1}^{n_t}$ of $n_t$ unlabeled samples. The source domain and target domain are sampled from joint distribution $P(\textbf{X}_s,\textbf{Y}_s)$ and $Q(\textbf{X}_t,\textbf{Y}_t)$, respectively, and $P \ne Q$. The goal of unsupervised domain adaptation is to learn discriminative features from source data and effectively transfer them from source to target to minimize target domain errors.


 There are two technical challenges to enabling successful domain adaptation: 1) promoting positive transfer of relevant discriminative features by enforcing cross-domain feature distribution consistency, and 2) reducing negative transfer of irrelevant features by preventing false across-domain feature alignment \cite{pan2010survey,pei2018multi}. Motivated by the two challenges, we propose to simultaneously enhance positive transfer by learning generic local feature patterns and alleviate negative transfer by enforcing additional local feature alignment. As illustrated in Figure 2, our method consists of three parts: \textbf{\Rmnum{1})} feature extractor, \textbf{\Rmnum{2})} local feature patterns learning, and \textbf{\Rmnum{3})} feature alignment. We employ multiple convolutional layers as the feature extractor to transform source and target data into feature maps with each position in the map representing a local feature. In the following sections, we describe how to learn local feature patterns and achieve feature alignment.

\subsection{Local Feature Patterns Learning}
In this section, we learn a cluster of discriminative local feature patterns to enable joint holistic and local feature alignment. We employ the end-to-end trainable NetVLAD for local feature patterns learning and local feature aggregation over the extracted convolutional feature maps \cite{arandjelovic2016netvlad}. We first learn a initial cluster of local feature patterns and then adapt them for cross-domain transfer. Given a collection of convolutional features from layer $\mathcal{L}_l$, we perform \emph{k-means} clustering to obtain the initial $K$ local feature patterns, $c_1^l,...,c_K^l$, represented by the $K$ clustering centers. For each image, a convolutional feature $F^l_{ij}$ at position $(i,j)$ of its feature map from layer $\mathcal{L}_l$ is assigned a similarity vector $S_{ij}^k$, defined as:

\begin{equation}
\small
S_{ij}^l[k]=\frac{e^{-\alpha{||F^l_{ij}-c_k^l||}^2}}{\Sigma_{k^{'}}{e^{-\alpha{||F^l_{ij}-c_{k^{'}}^l||}^2}}},
\end{equation}

\noindent which soft-assigns $F^l_{ij}$ to local feature pattern $c_k^l$ with weight proportional to its distances to the $k$ local feature patterns in the feature space. $S_{ij}^l[k]$ ranges between 0 and 1, with the highest \emph{similarity} value assigned to the closest local feature pattern. $\alpha$ is a tunable hyper-parameter (positive constant) and controls the decay of the similarity responses to the magnitude of the distances. Note that for $\alpha\rightarrow+\infty$, $F^l_{ij}$ is hard-assigned to the nearest local feature pattern. For the $d^l$ dimensional feature map $F^l$ from $\mathcal{L}_l$ , the NetVLAD encoding converts it into a single $d^l*K$ dimensional vector $V^l \in \mathbb{R}^{d^l*k}$, describing the distribution of local features regarding the $K$ local feature patterns. Formally, the encoding of an image regarding layer $\mathcal{L}_l$ is represented as:

\begin{equation}
\small
V^l[d,k]=\sum_{i=1}^{M^l}\sum_{j=1}^{N^l} S_{ij}^l[k] (F_{ij}^l[d]-c_{k}^l[d]),
\end{equation}

\noindent where $F_{ij}^l[d]$ and $c_{k}^l[d]$ are the $d$th dimension of feature $F_{ij}^l$ and local feature pattern $c_k^l$, respectively. $F_{ij}^l[d]-c_{k}^l[d]$ is the residual of feature $F_{ij}^l$ to local feature pattern $c_{k}^l$; ${M^l} \times {N^l}$ denotes the feature map size. The intuition is that residuals record the differences between the feature at each position and the typical local feature patterns. The residuals are aggregated inside each of the local feature pattern cell, and the similarity vector defined above determines the contribution of the residual of each feature to the total residuals. The output is a matrix $V$ with the $k$-th column $V[\cdot,k]$ representing the aggregated residuals inside the $k$-th local feature pattern cell. The columns of the matric are then stacked and normalized into a $d^l*K$ dimensional aggregated descriptor $V_h$ which is fed into the classifier for classification and holistic alignment.

We encourage the learned local feature patterns to be well-separated and local features to distribute compactly around them through minimizing a sparse loss $\mathcal{L}_{s}$ over the information entropy of the similarity vectors:

\begin{equation}
\small
\mathcal{L}_{s}=- \frac{1}{N_l*M_l} \sum_{i=1}^{M^l} \sum_{i=1}^{N^l} max(\sum_{k=1}^{K} S_{ij}^l[k] \log S_{ij}^l[k],m).
\end{equation}

\noindent where $m$ is the information entropy threshold. $S$ is the similarity vector described in Equation 1, but here we use a much smaller decay weight $\alpha_s$. Through the sparse loss minimization, we expect sparse soft-assignments of local features to the learned local feature patterns and less confusing boundary local features lying between different local feature pattern cells.

\subsection{Feature Alignment}
In this section, we aim to align source and target features based on the learned local feature patterns. We first describe adversarial learning and holistic alignment, and then present the additional local feature alignment.
\subsubsection{Adversarial Learning and Holistic Alignment}
 We employ the popular adversarial learning for distribution matching \cite{goodfellow2014generative,gulrajani2017improved}, as deep domain adversarial networks have achieved the top domain adaptation performances \cite{tzeng2017adversarial,pei2018multi}. The adversarial domain adaptation procedure is a two-player game, where the first player is the domain discriminator $\textbf{D}_h$ trained to distinguish source features from target features, while the second player, the feature extractor $\textbf{G}$, is trained to confuse the domain discriminator. By learning a best possible discriminator, the feature extractor is expected to learn features that are best domain-invariant. We achieve holistic alignment by matching the neural activation distributions of the classification layer. To be noted, the classification layer directly receives the aggregated local features over the learned local feature patterns. Formally, holistic domain discriminator the $\textbf{D}_h$ and feature extractor $\textbf{G}$ are trained to minimize loss $L_{D_h}$ and $L_{G_h}$, respectively, and they are defined as the following:

\begin{equation}
\small
\begin{split}
\mathcal{L}_{D_h}= -\frac{1}{n_s}\sum_{i=1}^{n_s}(log (\textbf{D}_h(\textbf{G}(x_s))))\\
       -\frac{1}{n_t}\sum_{i=1}^{n_t}(log (1-\textbf{D}_h(\textbf{G}(x_t)))),
\end{split}
\end{equation}

\begin{equation}
\small
\begin{split}
\mathcal{L}_{G_h}=-\frac{1}{n_t}\sum_{i=1}^{n_t}(log (\textbf{D}_h(\textbf{G}(x_t)))),
\end{split}
\end{equation}

\noindent where $n_s$ and $n_t$ are the number of training samples from source and target, respectively.

\subsubsection{Local Feature Alignment}
Existing adversarial domain adaptation methods only match the cross-domain holistic feature distributions and fail to consider the complex distributions of local features. As a result, multi-mode local feature patterns may be poorly aligned. Hence, we propose to further align the local features regarding the learned local feature patterns by minimizing an additional conditional domain adversarial loss \cite{mirza2014conditional,isola2017image}. For each convolutional feature $F_{ij}^l$ at position $(i,j)$ from layer $\mathcal{L}_l$, we hard-assign it to the nearest local feature pattern $a_{ij}^l$ , and $a_{ij}^l=\mathop{\arg\max}_{k}(S_{ij}^l[k])$. We align the residuals of local features to the assigned local feature patterns, and enforce that, within each local feature pattern cell, the residuals of local features from both domains distribute similarly. We find aligning the residuals, other than the original convolutional features, enables easier optimization and fine-grained alignment. Formally, an additional local feature discriminator $\textbf{D}_l$ is trained to minimize loss $L_{D_l}$ defined as:

\begin{equation}
\footnotesize
\begin{split}
\mathcal{L}_{D_l}= -\frac{1}{n_s}\sum_{n=1}^{n_s} \frac{1}{N_l*M_l}\sum_{i=1}^{M^l}\sum_{j=1}^{N^l}(log (\textbf{D}_l(F_{ij}^l-c^l[a_{ij}^l],a_{ij}^l))) \\
   -\frac{1}{n_t}\sum_{n=1}^{n_t} \frac{1}{N_l*M_l}\sum_{i=1}^{M^l}\sum_{j=1}^{N^l}(log (1-\textbf{D}_l(F_{ij}^l-c^l[a_{ij}^l],a_{ij}^l))),
\end{split}
\end{equation}

\noindent where $F_{ij}^l-c^l[a_{ij}^l]$ donates the residual of feature $F_{ij}^l$ to its assigned local feature pattern $a_{ij}^l$. For local feature alignment, the feature extractor network $\textbf{G}$ is trained to minimize loss $\mathcal{L}_{G_l}$ defined as:

\begin{equation}
\small
\begin{split}
\mathcal{L}_{G_l}=-\frac{1}{n_t}\sum_{n=1}^{n_t} \frac{1}{N_l*M_l}\sum_{i=1}^{M^l}\sum_{j=1}^{N^l}(log (\textbf{D}(F_{ij}^l-c^l[a_{ij}^l],a_{ij}^l))).
\end{split}
\end{equation}

To enable discriminative feature transferring, the feature extractor $\textbf{G}$ is also trained to minimize the classification loss $\mathcal{L}_c$ using source labels, defined as:

\begin{equation}
\small
\mathcal{L}_{c} = -\frac{1}{n_s}\sum_{i=1}^{n_s} y_{i}\cdot \log \hat{y}_{i},
\end{equation}

\noindent where $ y_{i}$ is the true label of the source sample $x_i$, and $\hat{y_{i}}$ is its predicted possibility.

Integrating all objectives together, the final loss for the feature extractor $\textbf{G}$ to minimize is

\begin{equation}
\small
\mathcal{L} = \mathcal{L}_{c}+\lambda_{h}\mathcal{L}_{G_h}+\lambda_{l}\mathcal{L}_{G_l}+\lambda_{s}\mathcal{L}_{s},
\end{equation}

\noindent where $\lambda_{h}$,  $\lambda_{l}$ and  $\lambda_{s}$ are hyper-parameter that trade-offs the objectives in the unified optimization problem. By optimizing the feature extractor network with the integrated loss, we aim to learn well-separated local feature patterns and simultaneously transfer category-related holistic features and generic local feature patterns.

\subsection{Implementation and Learning}

\subsubsection{Implementation Details} We use the VGG16 network \cite{simonyan2014very} as the backbone network and exploit the last convolutional layer, \emph{conv5\_3}, for local feature patterns learning and local alignment. We share the parameters of the source and target feature extractors. We append a single-layer classifier on the top of the aggregated local features. We keep the number of local feature patterns fixed to be 32. For local feature aggregation, we use a large $\alpha=5000.0$ to encourage independent residual accumulation within each local feature pattern cell. We use a small similarity decay $\alpha_s=0.005$ and a small sparsity threshold $m=0.02$. Since the dimensionality of the aggregated local features are large, $512*32$, we use a dropout of 0.5 over it to avoid over-fitting. For adversarial feature alignment, the holistic discriminator $\textbf{D}_h$ consists of 3 fully connected layers: two hidden layers with 768 and 1536 units, respectively, followed by the final discriminator output. We use larger 3-layer local adversarial discriminator $\textbf{D}_l$ with 2048 and 4096 units for the two hidden layers. We implement our model in Tensorflow and train it using Adam optimizer.

\subsubsection{Learning Procedure} We train our network in a three-step approach. In the first step, \emph{classifier training}, we initialize the local feature patterns using \emph{k-means} clustering, freeze them, and only train the one-layer classifier by minimizing the source classification loss with a learning rate of 0.01. In the second step, \emph{source finetuning}, we jointly finetune the classifier, local feature patterns, and the last two convolutional layers with a learning rate of 0.0001, and minimize the source classification loss combined with the sparsity loss. In the third step, \emph{domain adaptation}, we simultaneously train the classifier, local feature patterns, and the last two convolutional layers with a learning rate of 0.0001 to minimize the final joint loss described in Equation 9. Only finetuning and adapting the last two convolutional layers of VGG16 help to prevent overfitting to small datasets, reduce GPU memory footprint, and enable faster training. We set hyper-parameters $\lambda_{h}=0.2$,  $\lambda_{l}=0.1$ and  $\lambda_{s}=0.1$.

\section{Experiments}

We now evaluate our method with the state-of-the-art domain adaptation approaches on benchmark datasets. We experiment on the popular \emph{Office-31} dataset \cite{saenko2010adapting} and the recently introduced \emph{Office-home} dataset \cite{venkateswara2017deep}.

\noindent \textbf{Office-31} \cite{saenko2010adapting} This dataset is widely used for visual domain adaptation. It consists of 4,652 images and 31 categories collected from three different domains: Amazon (A), with 2817 images from amazon.com, Webcam (W) and DSLR (D), with 795 images and 498 images taken by web camera and digital SLR camera in different environmental settings, respectively. We evaluate all methods on the challenging settings of $A\leftrightarrow{W}$ and $A\leftrightarrow{D}$. The $W\leftrightarrow{D}$ performances are not reported as D and W are two similar domains and the domain shift is very small.

\noindent \textbf{Office-home} \cite{venkateswara2017deep} This is a very challenging domain adaptation dataset, which comprises 15,588 images with 65 categories of everyday objects in office and home settings. Some example samples are shown in Figure 3. There are 4 significantly different domains: Art (Ar) with 2427 painting, sketches or artistic depiction images, Clipart (Cl) with 4365 images, Product (Pr) containing 4439 images and Real-World (Rw) with 4357 regularly captured images. We report performances of all 12 transfer tasks to enable thorough evaluations: $Ar\leftrightarrow{Cl}$, $Ar\leftrightarrow{Pr}$, $Ar\leftrightarrow{Rw}$, $Cl\leftrightarrow{Pr}$, $Cl\leftrightarrow{Rw}$, and $Pr\leftrightarrow{Rw}$.

\begin{figure}[!hbtp]
    \centering
        \includegraphics[width=0.37 \textwidth]{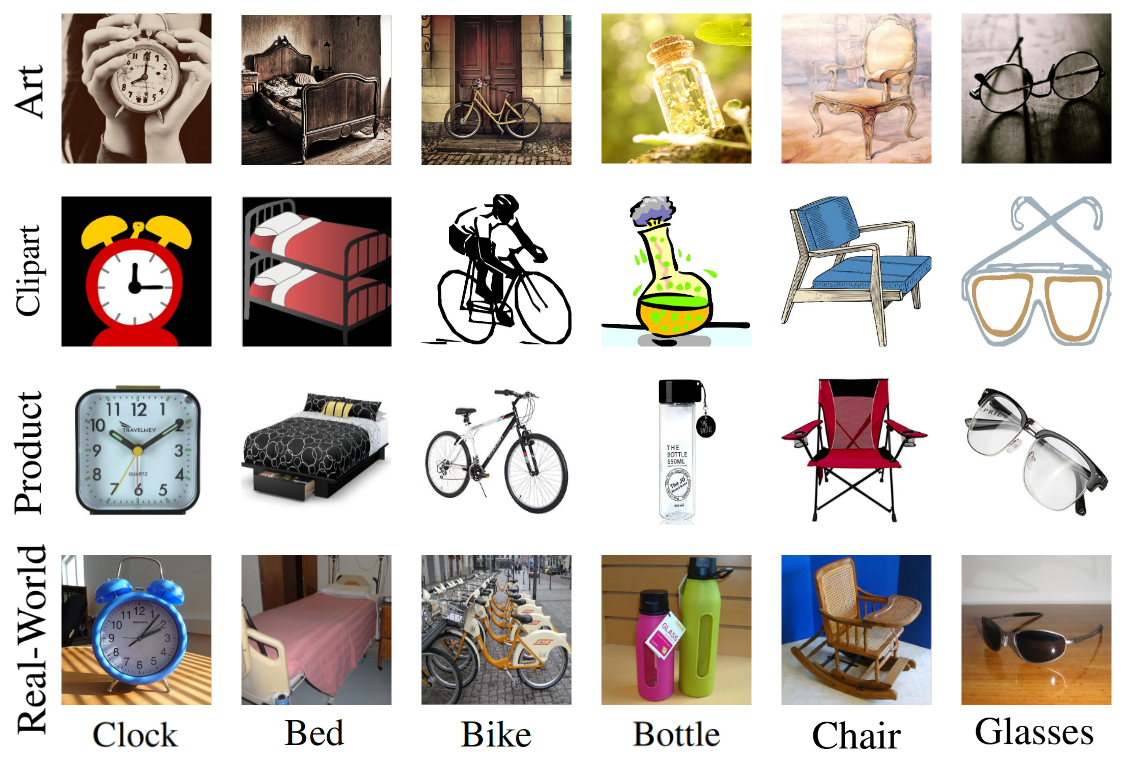}
   \caption{Example images of the \emph{Office-home} dataset.}
\end{figure}

\noindent \textbf{Compared Methods} We perform comparative studies of our method against the state-of-the-art deep domain adaptation methods: Deep CORAL (D-CORAL) \cite{sun2016deep}, Deep Adaptation Network (DAN) \cite{long2015learning}, Domain Adversarial Neural Network (DANN) \cite{ganin2016domain}, Adversarial Discriminative Domain Adaptation (ADDA) \cite{tzeng2017adversarial}, and Wasserstein Distance Guided Representation Learning (WD-GRL) \cite{shen2018wasserstein}. All these deep methods only align the holistic representations for domain adaptation. D-CORAL proposes to align the second-order statistics. DAN matches multi-layer deep features using multi-kernel MMD. DANN exploits adversarial learning for aligning deep features and enforces them indistinguishable for a additional domain discriminator. ADDA is a generalized framework for adversarial deep domain adaptation and unties weight sharing across domains. WD-GRL employs the Wasserstein distance to guide the adversarial learning of domain-invariant features.

 \noindent \textbf{Setup} We follow standard evaluation protocols for unsupervised domain adaptation: using all labeled source data and all unlabeled target data. We report the results averaged from three random experiments. The VGG16 network is used as the backbone model and the convolutional layers are initialized with parameters pre-trained on ImageNet dataset. The fully connected layers of our model are randomly initialized, while for comparing models they are initialized with parameters pre-trained on ImageNet dataset. To further explore the transferability of holistic features, we also report the performances of DAN and ADDA with fully-connected layers randomly initialized. In this case, to avoid overfitting, we use two smaller fully connected layers with 1024 and 128 hidden units, respectively, as done in \cite{motiian2017few}, and we denote them as DAN(s) and ADDA(s), respectively. To verify the importance of local feature alignment, we report results of our method in two different settings: 1) only holistic features are aligned ($\lambda_{h}=0.2,\lambda_{l}=0$), denoted as \emph{Our (H)}; 2) holistic features and local features are jointly aligned ($\lambda_{h}=0.2,\lambda_{l}=0.1$), denoted as \emph{Our (H+L)}.

\begin{table}[!htbp]
\small
  \centering
  \caption{Accuracy ($\%$) on the \emph{Office31} dataset for unsupervised domain adaptation.}
    \begin{tabular}{cccccc}
    \toprule[1pt]
    Method &A$\rightarrow$W&A$\rightarrow$D&W$\rightarrow$A&D$\rightarrow$A &Avg\\
    \midrule
    D-CORAL&$69.23$&$68.52$&$59.53$&$60.53$&$64.45$\\
    DAN&$70.35$&$69.56$&$61.38$&$60.92$&$65.55$\\
    WD-GRL &$74.61$&$73.27$&$60.65$&$60.01$&$67.14$\\
    ADDA&$75.24$&$73.64$&$60.34$&$60.75$&$67.49$\\
    DANN&$74.03$&$73.52$&$62.23$&$61.73$&$67.88$\\
    \midrule
    ADDA(s)&$64.89$&$64.84$&$48.66$&$50.07$&$57.12$\\
     DANN(s)&$64.85$&$61.48$&$40.50$&$48.21$&$53.76$\\
    \midrule
    Ours(H)&$82.63$&$75.02$&$63.47$&$\textbf{64.16}$&$71.32$\\
    Ours(H+L)&$\textbf{84.35}$&$\textbf{77.56}$&$\textbf{64.56}$&$63.38$&$\textbf{72.46}$\\
    \bottomrule[1pt]
    \end{tabular}%
\end{table}%

\begin{table*}[!htbp]
 \small
  \centering
  \caption{Accuracy ($\%$) on the \emph{Office-home} dataset for unsupervised domain adaptation.}
  \resizebox{\textwidth}{20mm}{
    \begin{tabular}{cccccccccccccc}
    \toprule[1pt]
    Method &Ar$\rightarrow$Cl&Ar$\rightarrow$Pr&Ar$\rightarrow$Rw&Cl$\rightarrow$Ar&Cl$\rightarrow$Pr&Cl$\rightarrow$Rw&Pr$\rightarrow$Ar&Pr$\rightarrow$Cl&Pr$\rightarrow$Rw&Rw$\rightarrow$Ar &Rw$\rightarrow$Cl&Rw$\rightarrow$Pr&Avg\\
    \midrule
    D-CORAL&$35.85$&$45.58$&$56.52$&$36.28$&$47.54$&$50.27$&$36.45$&$37.57$&$61.54$&$49.87$&$43.21$&$69.01$&$47.47$\\
     DAN    &$36.98$&$45.71$&$58.11$&$38.90$&$50.33$&$51.94$&$37.44$&$37.05$&$61.49$&$50.07$&$42.18$&$68.37$&$48.21$\\
     WD-GRL    &$36.37$&$44.54$&$58.12$&$37.65$&$51.54$&$52.46$&$36.73$&$37.98$&$63.08$&$50.04$&$44.49$&$70.15$&$48.60$\\
     ADDA   &$38.33$&$47.81$&$60.54$&$36.34$&$49.56$&$50.54$&$37.89$&$39.26$&$64.75$&$54.14$&$46.55$&$71.64$&$49.78$\\
     DANN   &$37.04$&$46.29$&$59.38$&$39.83$&$52.95$&$53.85$&$37.80$&$38.52$&$63.86$&$49.61$&$45.15$&$70.76$&$49.59$\\
    \midrule
    ADDA(s)&$25.05$&$35.30$&$42.23$&$27.71$&$37.37$&$39.98$&$30.15$&$34.94$&$55.34$&$45.30$&$39.50$&$62.20$&$39.59$\\
     DANN(s)&$23.16$&$33.75$&$41.91$&$25.96$&$38.72$&$37.73$&$29.20$&$35.00$&$50.80$&$42.02$&$38.19$&$61.03$&$38.12$\\
    \midrule
    Ours(H)&$38.95$&$48.05$&$62.33$&$40.56$&$53.45$&$56.36$&$37.85$&$39.43$&$65.18$&$54.78$&$46.29$&$75.21$&$51.54$\\
    Ours(H+L)&$\textbf{41.53}$&$\textbf{53.66}$&$\textbf{64.90}$&$\textbf{41.53}$&$\textbf{54.57}$&$\textbf{57.66}$&$\textbf{38.87}$&$\textbf{40.08}$&$\textbf{65.97}$&$\textbf{55.13}$&$\textbf{47.18}$&$\textbf{76.02}$&$\textbf{53.10}$\\
    \bottomrule[1pt]
    \end{tabular}}%
\end{table*}%

\subsection{Results}

The results on the \emph{Office-31} dataset are shown in Table 1. The proposed method outperforms all the compared models, though we do not use any pre-trained fully-connected layers. The average improvements of \emph{Our (H+L)} over ADDA and DANN are $4.97\%$ and $4.58\%$, respectively. The obvious advantages of \emph{Our (H)} over the compared holistic feature based models verify the superior transferability of the learned local feature patters, as all of them enforce the similar holistic alignment. We observe that the improvements of our method are more obvious when \emph{A} acts as the source domain. Domain \emph{A} comprises more training images with more diversities, and thus more generic local feature patterns can be learned, which effectively enhance positive feature transfer.

The performances of ADDA and DANN drop significantly when the fully-connected layers are trained from scratch using the source labels, and the averaged performance gaps from the pretrained models are $10.37\%$ and $14.12\%$, respectively. The performance drops are more distinct when \emph{D} or \emph{W} acts as the source domain, as these two domains have much less training images and the learned holistic representations severely overfit the source labels. The results verify the inferior transferability of holistic representations, especially when source labels are limited.

The performances of all methods on the \emph{Office-home} dataset are reported in Table 2. The proposed model \emph{Our (H+L)} achieves consistent improvements over the comparison methods. For \emph{Office-home} dataset, the training images in each category show more diversities as verified by the lower in-domain classification accuracy described in its original paper \cite{venkateswara2017deep}. The model \emph{Our (H+L)} shows consistent advantages over the model \emph{Our (H)}, and the advantages are manifested in settings when \emph{Art} acts as the source domain. Adaptations from \emph{Art} to other domains are more challenging as images from \emph{Art} show more diversities within each category while having nearly half of the training samples of the other three domains. That means more complex local feature patterns with less referable points (holistic features) to be transferred from \emph{Art} source domain. In this case, enforcing additional local feature alignment promotes positive transfer of relevant local features within each local feature pattern cell, and thus improves performances.

\textbf{Negative transfer} happens when features are falsely aligned and domain adaptation causes deteriorated performances. Existing holistic feature distribution matching easily induce negative transfer when the distributions between source and target are inherently different. Consider the case when source domain is much smaller or larger than target. We aim to evaluate the robustness of domain adaptation methods against negative transfer in a more common scenario where source domain is much larger than target. In this case, there are many source points in the feature space (semantic features) that are irrelevant to the target domain. We experiment with setting, \emph{31-25}, on the four transfer tasks constructed from the \emph{Office-31} dataset, by removing the last 6 classes in alphabet order from the target domain. For example, we perform domain adaptation on transfer task \emph{A31-D25}, where the source domain \emph{A} has 31 classes but the target domain \emph{W} only contains 25 classes. The results are reported in Table 3. As we can see, there are obvious negative transfer for top-performing domain adaptation methods DANN and ADDA. Both of them under-perform the finetuned VGG16 net on most of the transfer tasks and the averaged performance drops are $2.59\%$ and $10.34\%$, respectively. Our domain adaptation model \emph{Our (H+L)} and \emph{Our (H)} both outperform the unadapted model \emph{Our(w/o DA)}. The significant improvements bringed by our domain adaptation method prove the advantage of exploiting generic local feature patterns in combating negative transfer.

\begin{table}[!htbp]
  \centering \small
  \caption{Accuracy ($\%$) on the \emph{Office31} dataset for unsupervised domain adaptation from 31 to 25 categories.}
    \begin{tabular}{cccccc}
    \toprule[1pt]
    Method &A$\rightarrow$W&A$\rightarrow$D&W$\rightarrow$A&D$\rightarrow$A &Avg\\
    \midrule
    VGG16(FT)&$63.33$&$65.92$&$58.68$&$57.83$&$61.44$\\
    ADDA&$63.21$&$59.79$&$44.74$&$36.65$&$51.10$\\
    DANN&$62.12$&$62.31$&$59.46$&$51.49$&$58.85$\\
    \midrule
    Our(w/o DA) &$62.51$&$63.67$&$56.47$&$56.35$&$59.75$\\
    Ours(H)&$74.85$&$68.77$&$63.88$&$62.82$&$67.58$\\
    Ours(H+L)&$\textbf{75.49}$&$\textbf{70.05}$&$\textbf{67.67}$&$\textbf{63.14}$&$\textbf{69.09}$\\
    \bottomrule[1pt]
    \end{tabular}%
\end{table}%

\begin{figure*}
\centering
\subcaptionbox{VGG16 finetuned }{\includegraphics[width=0.15\textwidth]{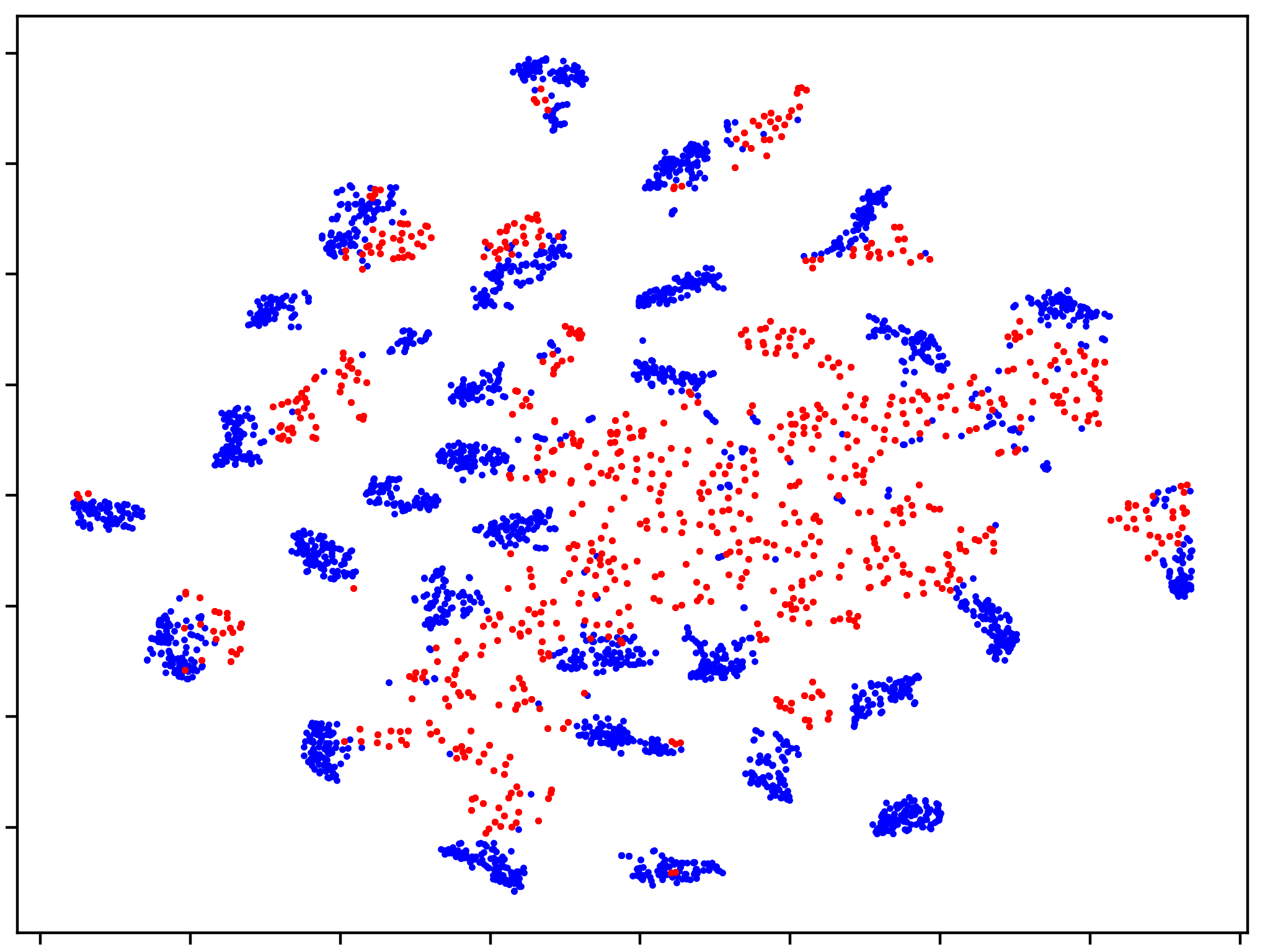}}%
\hfill
\subcaptionbox{DANN}{\includegraphics[width=0.15\textwidth]{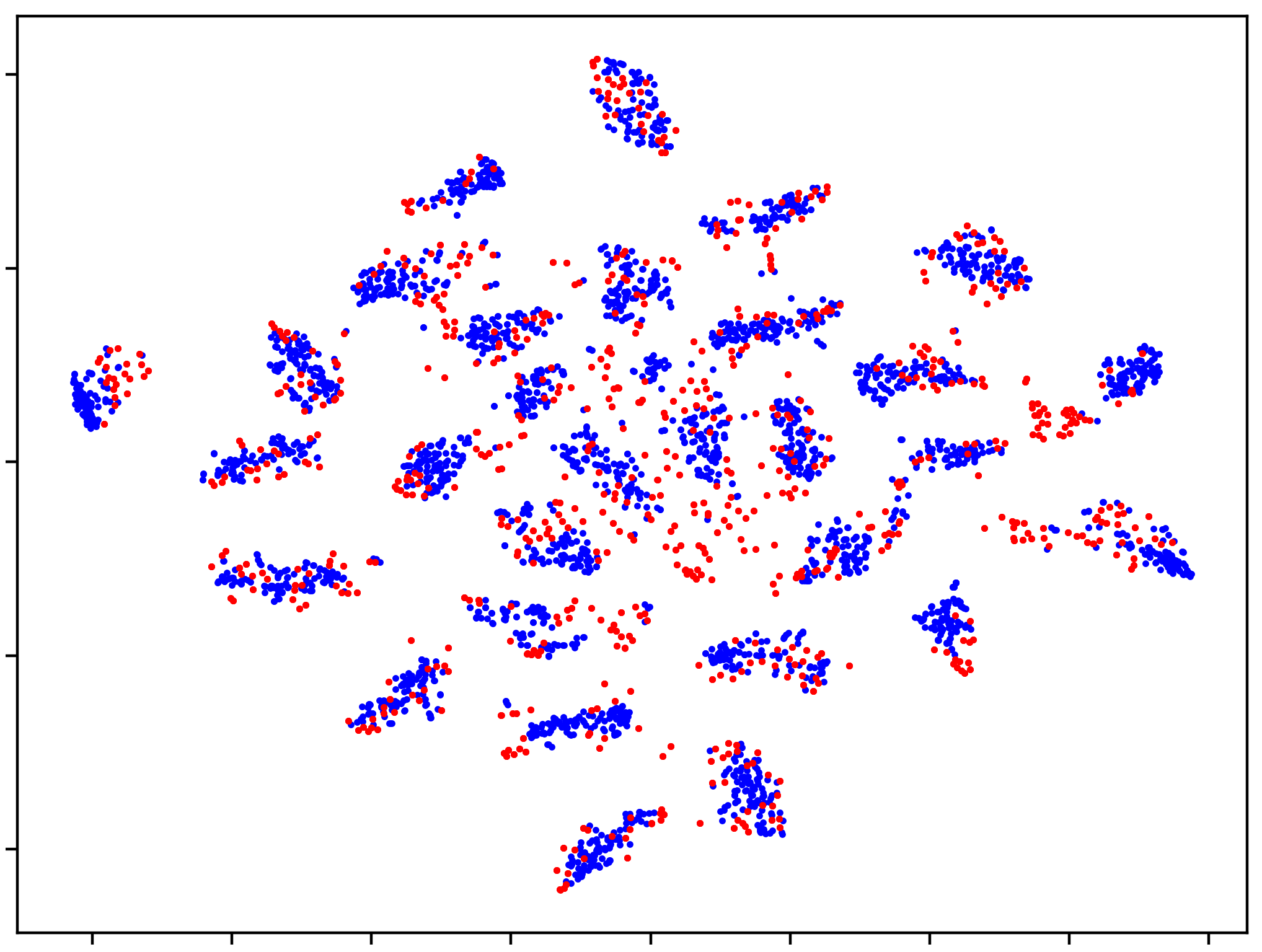}}%
\hfill
\subcaptionbox{ Our (unadapted)}{\includegraphics[width=0.15\textwidth]{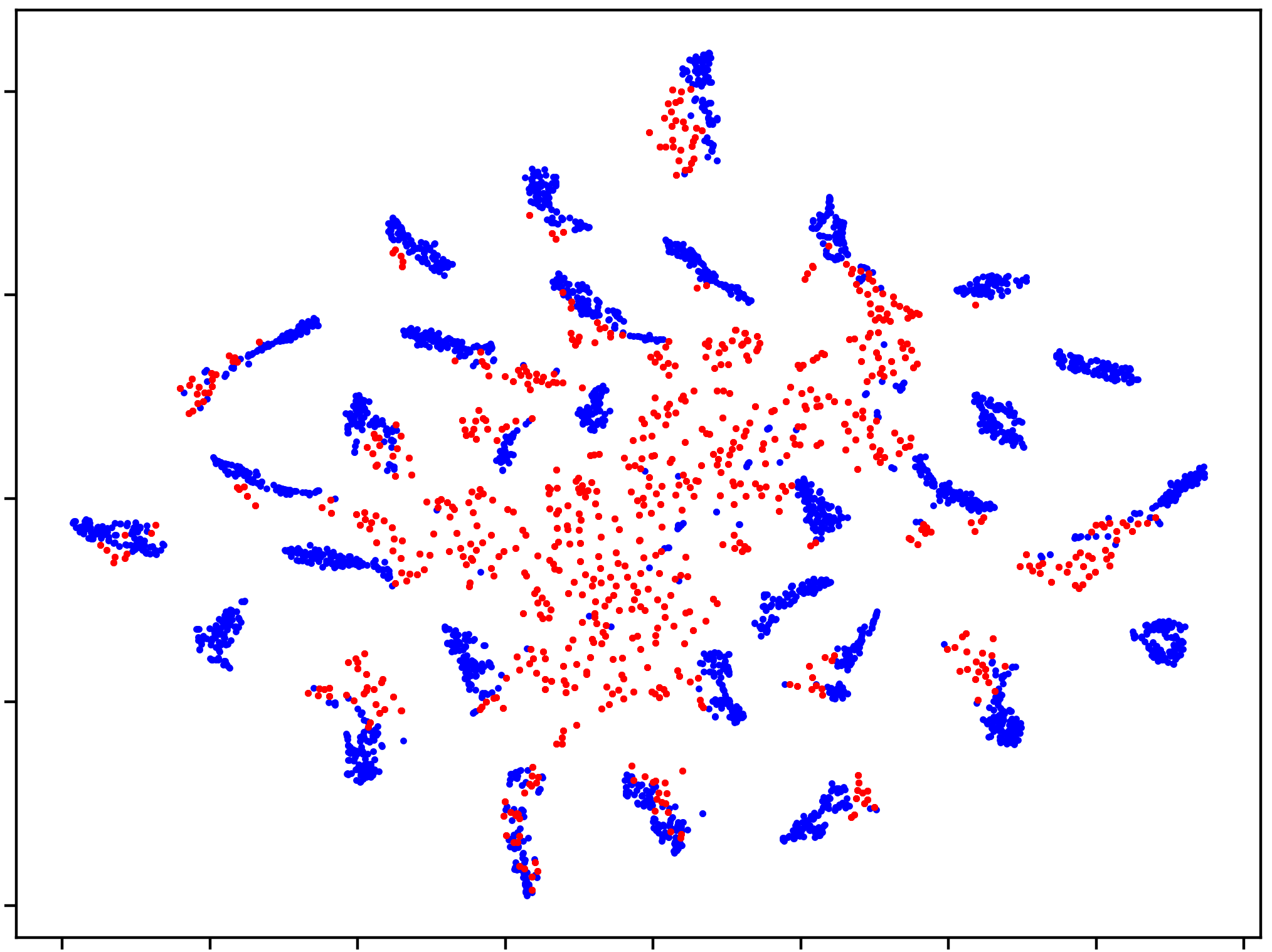}}
\hfill
\subcaptionbox{\emph{Our (H+L)}}{\includegraphics[width=0.15\textwidth]{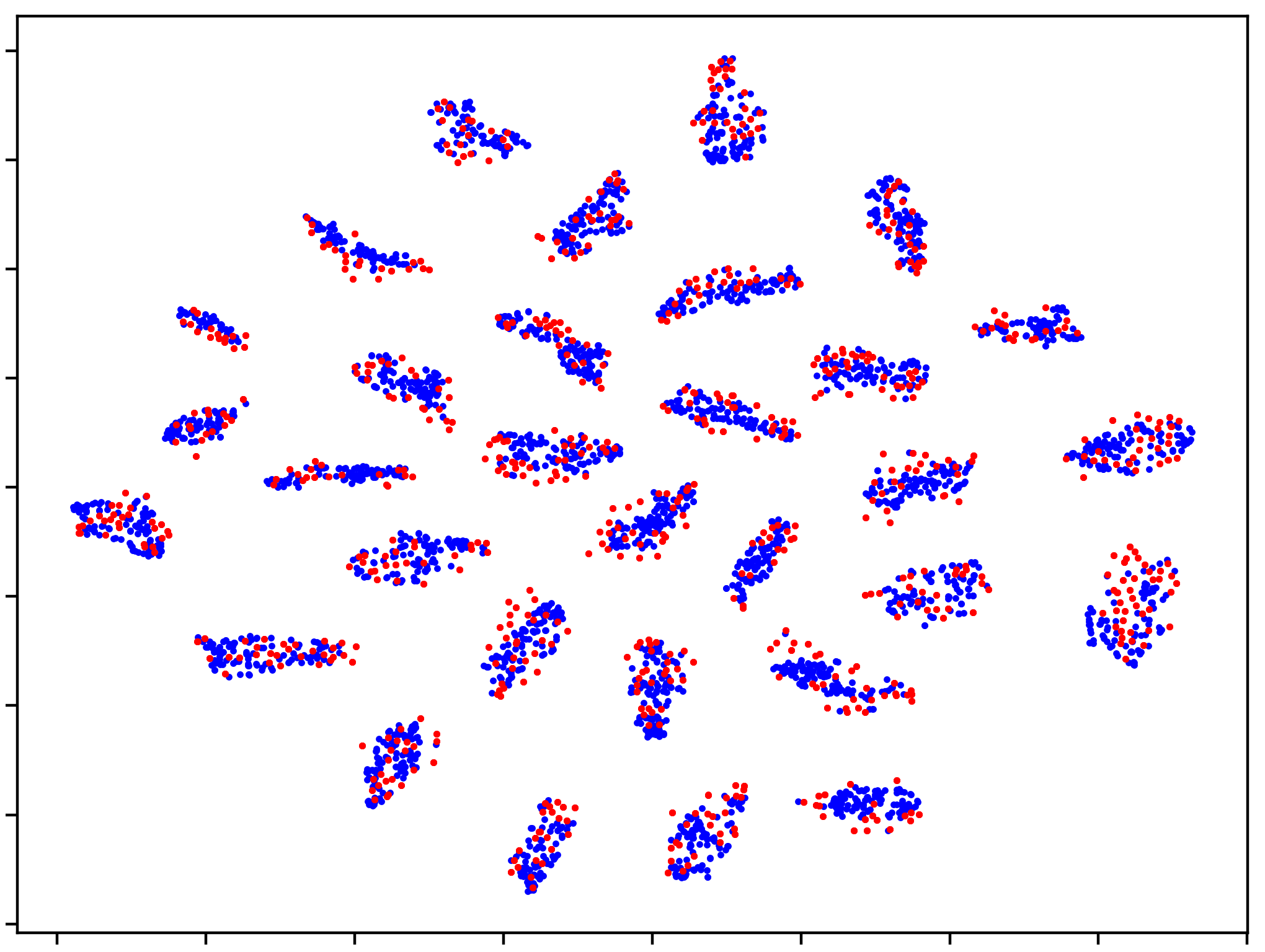}}%
\caption{The t-SNE visualizations of holistic representations learned by (a) Fine-tuned VGG16 net, (b) DANN, (c) Our unadapted model, and (d) Adapted \emph{Our (H+L)} (blue: \textbf{A}, red: \textbf{W}).}
\end{figure*}

\begin{figure*}
\centering
\subcaptionbox{VGG16}{\includegraphics[width=0.15\textwidth]{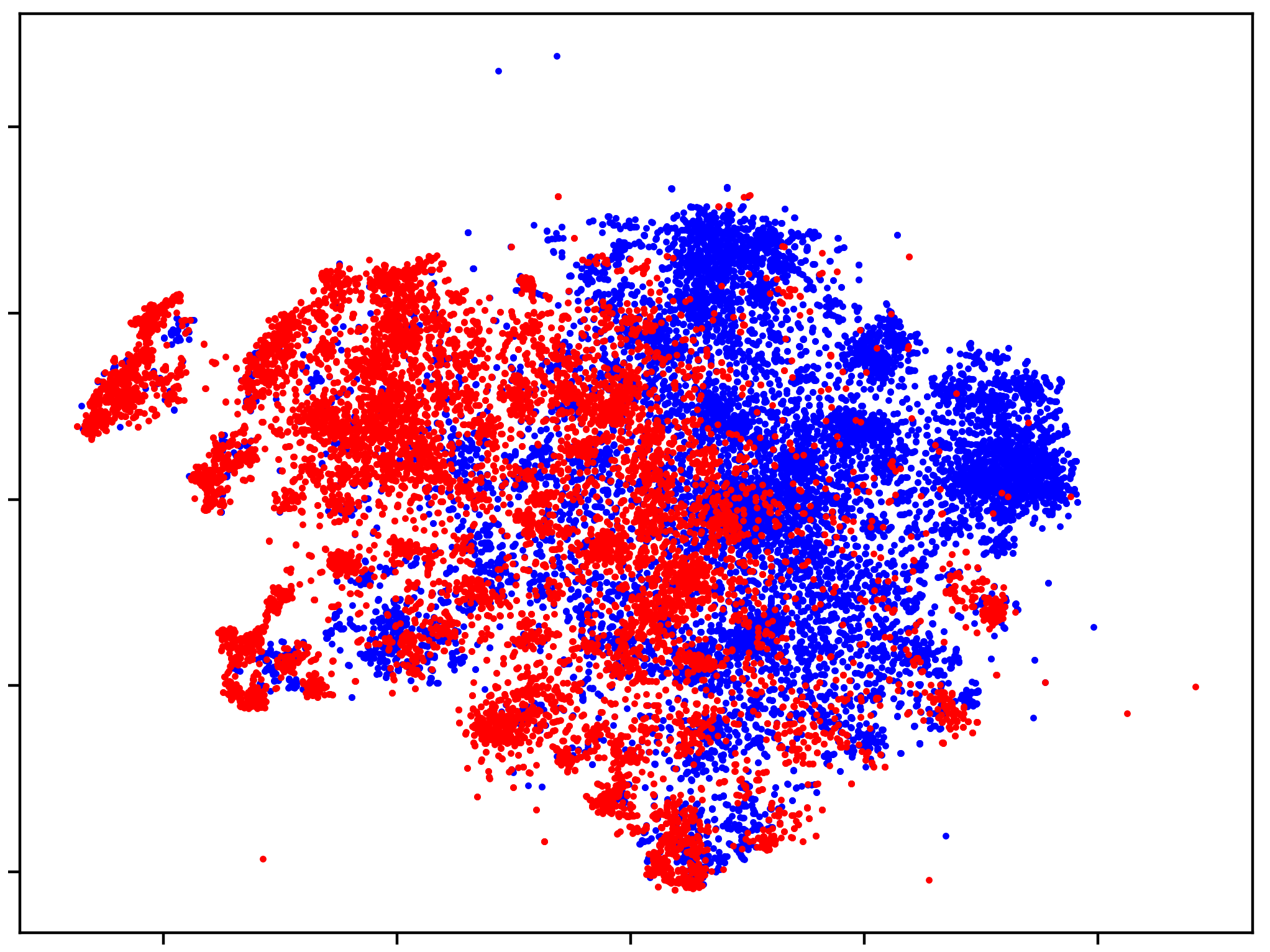}}%
\hfill
\subcaptionbox{DANN}{\includegraphics[width=0.15\textwidth]{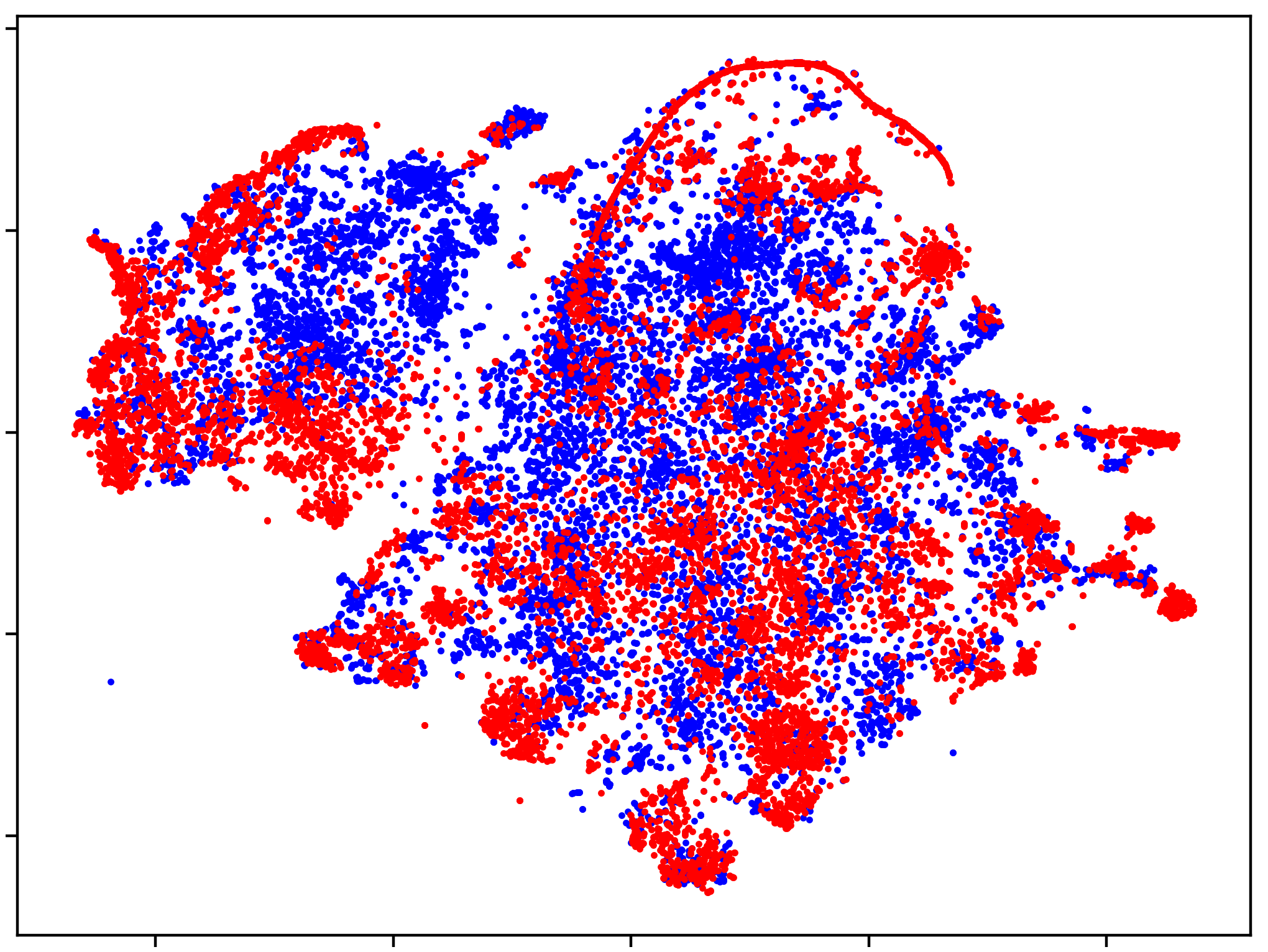}}
\hfill
\subcaptionbox{Our (H)}{\includegraphics[width=0.15\textwidth]{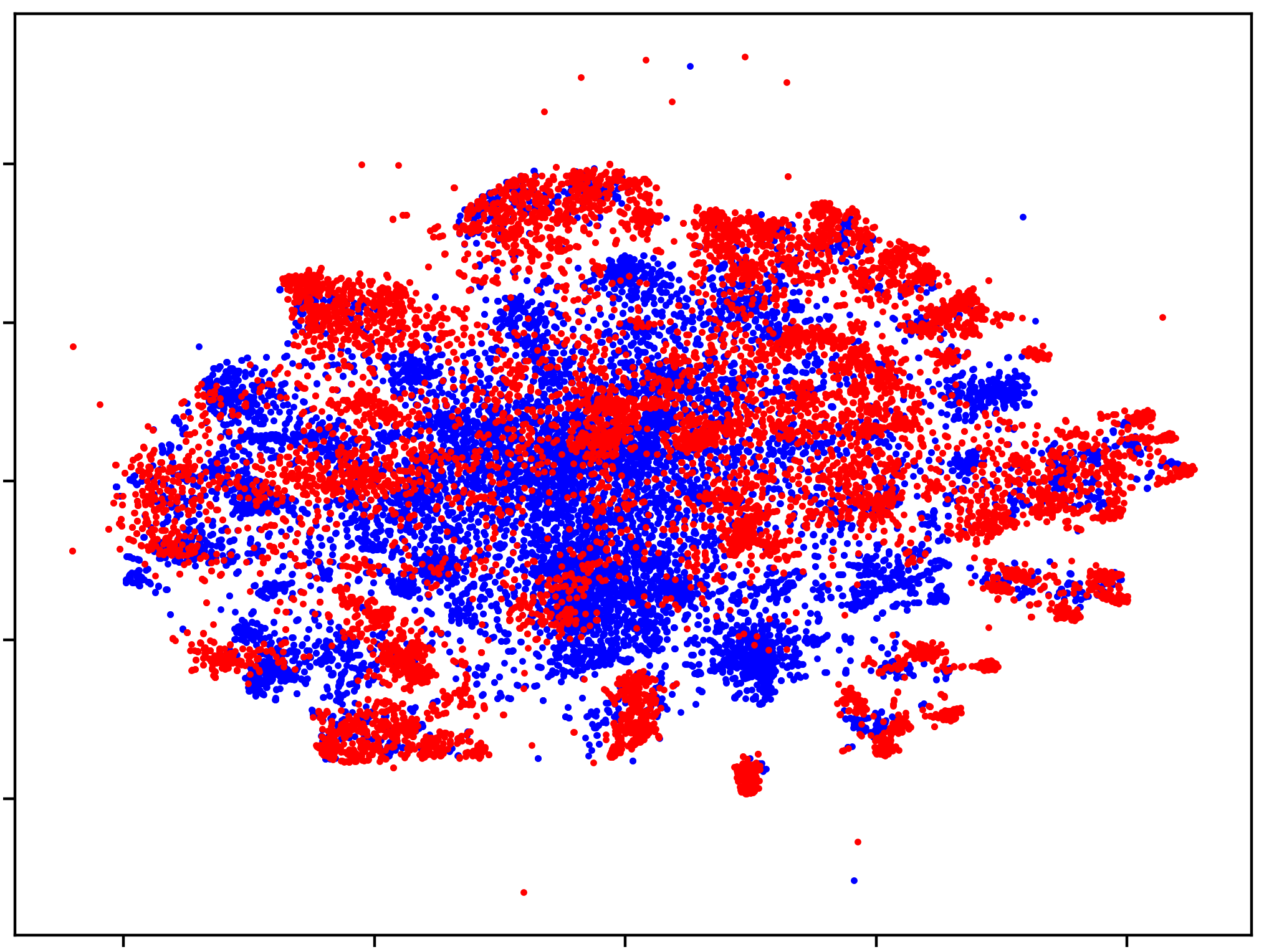}}%
\hfill
\subcaptionbox{Our (H+L)}{\includegraphics[width=0.15\textwidth]{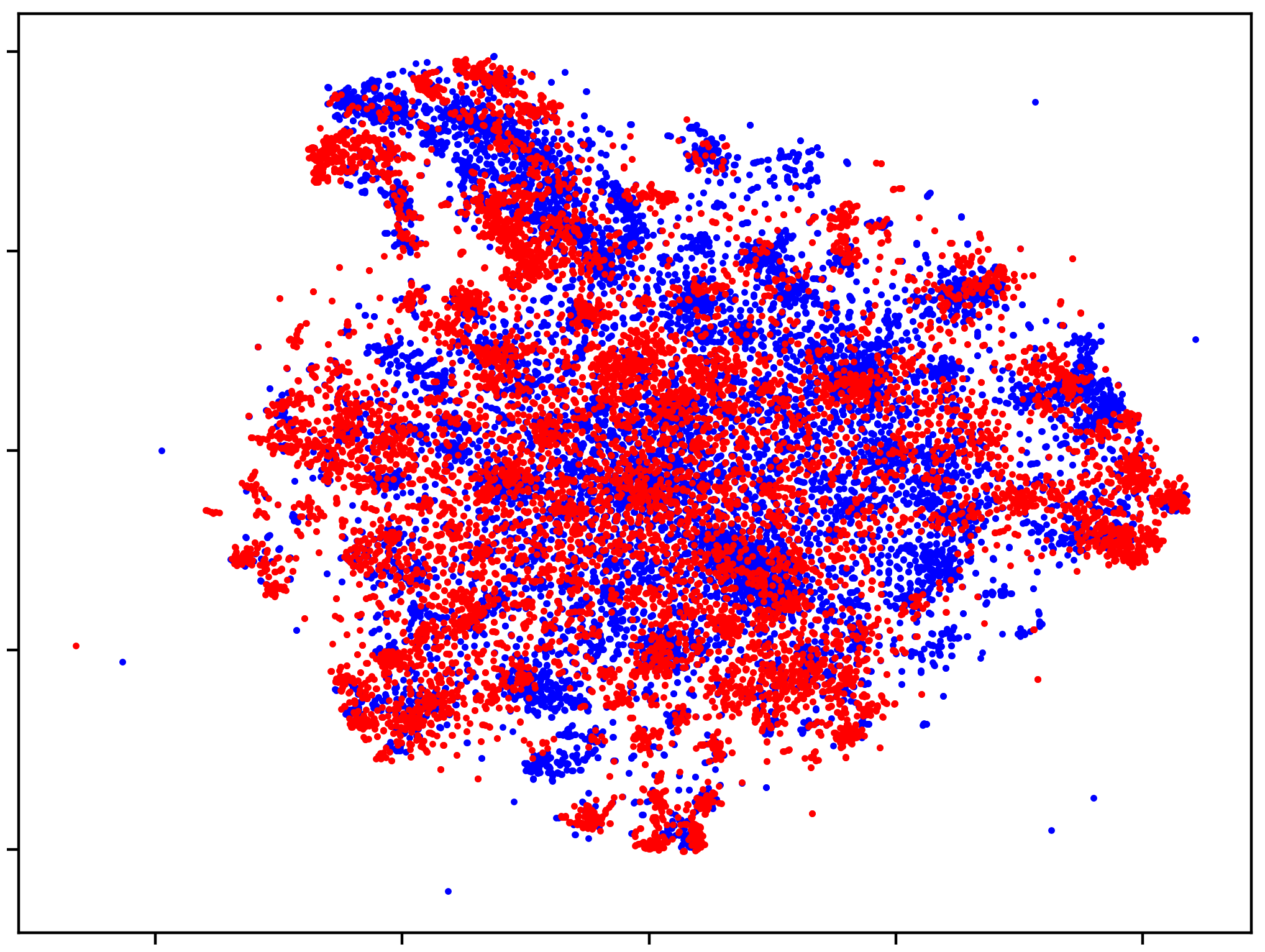}}%
\caption{The t-SNE visualizations of local features, \emph{conv5\_3}, of (a) VGG16 net, (b) DANN, (c) \emph{Our (H)}, and (d) \emph{Our (H+L)} (blue: \textbf{A}, red: \textbf{W}).}
\end{figure*}

\subsection{Ablation Study}
\noindent \textbf{Comparison of Layers} We compare the performances of models trained and adapted from different convolutional layers, and the results are shown in Table 4. As we can see, the performance trends are clear: higher layers achieve better performances and the best performance is achieved by layer \emph{conv5\_3}. Lower layers are more generic, however, much less semantic features are captured if the higher layer features are abandoned.

\begin{table}[!htbp]
\small
  \centering
  \caption{Accuracy ($\%$) on the \emph{Office31} dataset with convolutional features from different layers.}
    \begin{tabular}{cccccc}
    \toprule[1pt]
    Layers &A$\rightarrow$W&A$\rightarrow$D&W$\rightarrow$A&D$\rightarrow$A &Avg\\
    \midrule
    conv5\_1 &$73.70$&$70.01$&$43.96$&$39.88$&$56.89$\\
    conv5\_2 &$79.62$&$73.49$&$50.96$&$50.60$&$63.67$\\
    conv5\_3 &$\textbf{84.35}$&$\textbf{77.56}$&$\textbf{64.56}$&$\textbf{63.38}$&$\textbf{72.46}$\\
    \bottomrule[1pt]
    \end{tabular}%
\end{table}%

\noindent \textbf{Number of Local Feature Patterns} We explore the effects of the size of local feature patterns on the \emph{Office-31} dataset. We report the average accuracy of the four transfer tasks in Table 5. As we can see, the performances are non-sensitive to the size of local feature patterns. With larger sizes, we achieve improved performances. Larger sizes of local feature patterns mean more complex local distributions, hence we need more powerful local discriminator $\textbf{D}_l$ to distinguish source from target. For example, for $k=64$, the hidden units number of the $\textbf{D}_l$ are both 4096 for the two layers.

\begin{table}[!htbp]
\small
  \centering
  \caption{Accuracy ($\%$ ) on the \emph{Office31} dataset with varying size of local feature patterns.}
    \begin{tabular}{cccccc}
    \toprule[1pt]
    Nunmber &k=0&k=8&k=16&k=32&k=64\\
    \midrule
    Accuracy &$67.92$&$69.75$&$70.84$&$72.46$&$\textbf{72.59}$\\
    \bottomrule[1pt]
    \end{tabular}%
\end{table}%

\noindent \textbf{Alignment Visualization} We visualize both the holistic representations and local features using t-SNE embedding with the $A\rightarrow{W}$ transfer task. In Figure 4, we visualize the network activations of the last fully-connected layer of finetuned VGG16 net, DANN, our unadapted model, and our adapted model \emph{Our (H+L)}. For both the finetuned VGG16 and our unadapted model, target and source are poorly aligned. For DANN, as shown in Figure 4(b), source and target representations are well aligned, but there are still many boundary confusing points lying between different category clusters. For our adapted model \emph{Our (H+L)}, source and target representations are much better aligned.

In Figure 5, we visualize the network activations of the last convolutional layer, \emph{conv5\_3}, to study the effects of local alignment. As shown in Figure 5 (a), source and target local features are poorly aligned for the VGG16 net.  When adapted with DANN and \emph{Our (H)}, local features are better aligned, though both of the two models only match holistic feature distributions. As our model encourages learning well-separated local feature patterns by minimizing an additional sparsity loss, the local features learned by our model distribute in more compact clusters (best view the zoomed-in figure). When enforcing additional local alignment by \emph{Our (H+L)}, local feature patterns tend to be equally shared by source and target, and local features are better aligned within each local feature pattern cluster.

\section{ Conclusions}
We have presented a novel and effective approach to exploiting local features for unsupervised domain adaptation. Unlike existing deep domain adaptation methods that only transfer holistic representations, the proposed method learns domain-invariant local feature patterns, and simultaneously aligns holistic features and local features to enable fine-grained feature alignment. Experimental results verified the advantages of the proposed method over the state-of-the-art unsupervised domain adaptation approaches. We have explored the performances of convolutional features from different layers for domain adaptation with the VGG16 net and found that the last convolutional layer achieves the best performances. Further, we showed that the proposed method can effectively alleviate negative transfer.

\section{ Acknowledgments}
This work is supported by the Zhejiang Provincial Natural Science Foundation (LR19F020005), National Natural Science Foundation of China (61572433, 61672125, 31471063, 61473259, 31671074) and thanks for a gift grant from Baidu inc. We are also partially supported by the Hunan Provincial Science and Technology Project Foundation (2018TP1018, 2018RS3065) and the Fundamental Research Funds for the Central Universities.

\bibliographystyle{aaai}
\bibliography{bibfilefile}

\begin{thebibliography}{}

\bibitem[\protect\citeauthoryear{Arandjelovic \bgroup et al\mbox.\egroup
  }{2016}]{arandjelovic2016netvlad}
Arandjelovic, R.; Gronat, P.; Torii, A.; Pajdla, T.; and Sivic, J.
\newblock 2016.
\newblock Netvlad: Cnn architecture for weakly supervised place recognition.
\newblock In {\em Proceedings of the IEEE Conference on Computer Vision and
  Pattern Recognition},  5297--5307.

\bibitem[\protect\citeauthoryear{Ben-David \bgroup et al\mbox.\egroup
  }{2010}]{ben2010theory}
Ben-David, S.; Blitzer, J.; Crammer, K.; Kulesza, A.; Pereira, F.; and Vaughan,
  J.~W.
\newblock 2010.
\newblock A theory of learning from different domains.
\newblock {\em Machine learning} 79(1-2):151--175.

\bibitem[\protect\citeauthoryear{Bengio, Courville, and
  Vincent}{2013}]{bengio2013representation}
Bengio, Y.; Courville, A.; and Vincent, P.
\newblock 2013.
\newblock Representation learning: A review and new perspectives.
\newblock {\em IEEE transactions on pattern analysis and machine intelligence}
  35(8):1798--1828.

\bibitem[\protect\citeauthoryear{Csurka}{2017}]{csurka2017domain}
Csurka, G.
\newblock 2017.
\newblock Domain adaptation for visual applications: A comprehensive survey.
\newblock {\em arXiv preprint arXiv:1702.05374}.

\bibitem[\protect\citeauthoryear{Fernando \bgroup et al\mbox.\egroup
  }{2013}]{fernando2013unsupervised}
Fernando, B.; Habrard, A.; Sebban, M.; and Tuytelaars, T.
\newblock 2013.
\newblock Unsupervised visual domain adaptation using subspace alignment.
\newblock In {\em Proceedings of the IEEE international conference on computer
  vision},  2960--2967.

\bibitem[\protect\citeauthoryear{Ganin \bgroup et al\mbox.\egroup
  }{2016}]{ganin2016domain}
Ganin, Y.; Ustinova, E.; Ajakan, H.; Germain, P.; Larochelle, H.; Laviolette,
  F.; Marchand, M.; and Lempitsky, V.
\newblock 2016.
\newblock Domain-adversarial training of neural networks.
\newblock {\em The Journal of Machine Learning Research} 17(1):2096--2030.

\bibitem[\protect\citeauthoryear{Ghifary \bgroup et al\mbox.\egroup
  }{2016}]{ghifary2016deep}
Ghifary, M.; Kleijn, W.~B.; Zhang, M.; Balduzzi, D.; and Li, W.
\newblock 2016.
\newblock Deep reconstruction-classification networks for unsupervised domain
  adaptation.
\newblock In {\em European Conference on Computer Vision},  597--613.
\newblock Springer.

\bibitem[\protect\citeauthoryear{Girdhar \bgroup et al\mbox.\egroup
  }{2017}]{girdhar2017actionvlad}
Girdhar, R.; Ramanan, D.; Gupta, A.; Sivic, J.; and Russell, B.
\newblock 2017.
\newblock Actionvlad: Learning spatio-temporal aggregation for action
  classification.
\newblock In {\em CVPR}, volume~2, ~3.

\bibitem[\protect\citeauthoryear{Gong \bgroup et al\mbox.\egroup
  }{2012}]{gong2012geodesic}
Gong, B.; Shi, Y.; Sha, F.; and Grauman, K.
\newblock 2012.
\newblock Geodesic flow kernel for unsupervised domain adaptation.
\newblock In {\em Computer Vision and Pattern Recognition (CVPR), 2012 IEEE
  Conference on},  2066--2073.
\newblock IEEE.

\bibitem[\protect\citeauthoryear{Goodfellow \bgroup et al\mbox.\egroup
  }{2014}]{goodfellow2014generative}
Goodfellow, I.; Pouget-Abadie, J.; Mirza, M.; Xu, B.; Warde-Farley, D.; Ozair,
  S.; Courville, A.; and Bengio, Y.
\newblock 2014.
\newblock Generative adversarial nets.
\newblock In {\em Advances in neural information processing systems},
  2672--2680.

\bibitem[\protect\citeauthoryear{Gulrajani \bgroup et al\mbox.\egroup
  }{2017}]{gulrajani2017improved}
Gulrajani, I.; Ahmed, F.; Arjovsky, M.; Dumoulin, V.; and Courville, A.~C.
\newblock 2017.
\newblock Improved training of wasserstein gans.
\newblock In {\em Advances in Neural Information Processing Systems},
  5767--5777.

\bibitem[\protect\citeauthoryear{Hong \bgroup et al\mbox.\egroup
  }{2018}]{hong2018conditional}
Hong, W.; Wang, Z.; Yang, M.; and Yuan, J.
\newblock 2018.
\newblock Conditional generative adversarial network for structured domain
  adaptation.
\newblock In {\em Proceedings of the IEEE Conference on Computer Vision and
  Pattern Recognition},  1335--1344.

\bibitem[\protect\citeauthoryear{Isola \bgroup et al\mbox.\egroup
  }{2017}]{isola2017image}
Isola, P.; Zhu, J.-Y.; Zhou, T.; and Efros, A.~A.
\newblock 2017.
\newblock Image-to-image translation with conditional adversarial networks.
\newblock In {\em Proceedings of the IEEE Conference on Computer Vision and
  Pattern Recognition},  1125--1134.

\bibitem[\protect\citeauthoryear{J{\'e}gou \bgroup et al\mbox.\egroup
  }{2010}]{jegou2010aggregating}
J{\'e}gou, H.; Douze, M.; Schmid, C.; and P{\'e}rez, P.
\newblock 2010.
\newblock Aggregating local descriptors into a compact image representation.
\newblock In {\em Computer Vision and Pattern Recognition (CVPR), 2010 IEEE
  Conference on},  3304--3311.
\newblock IEEE.

\bibitem[\protect\citeauthoryear{Long \bgroup et al\mbox.\egroup
  }{2015}]{long2015learning}
Long, M.; Cao, Y.; Wang, J.; and Jordan, M.~I.
\newblock 2015.
\newblock Learning transferable features with deep adaptation networks.
\newblock {\em arXiv preprint arXiv:1502.02791}.

\bibitem[\protect\citeauthoryear{Mirza and
  Osindero}{2014}]{mirza2014conditional}
Mirza, M., and Osindero, S.
\newblock 2014.
\newblock Conditional generative adversarial nets.
\newblock {\em arXiv preprint arXiv:1411.1784}.

\bibitem[\protect\citeauthoryear{Motiian \bgroup et al\mbox.\egroup
  }{2017}]{motiian2017few}
Motiian, S.; Jones, Q.; Iranmanesh, S.; and Doretto, G.
\newblock 2017.
\newblock Few-shot adversarial domain adaptation.
\newblock In {\em Advances in Neural Information Processing Systems},
  6670--6680.

\bibitem[\protect\citeauthoryear{Pan \bgroup et al\mbox.\egroup
  }{2011}]{pan2011domain}
Pan, S.~J.; Tsang, I.~W.; Kwok, J.~T.; and Yang, Q.
\newblock 2011.
\newblock Domain adaptation via transfer component analysis.
\newblock {\em IEEE Transactions on Neural Networks} 22(2):199--210.

\bibitem[\protect\citeauthoryear{Pan, Yang, and others}{2010}]{pan2010survey}
Pan, S.~J.; Yang, Q.; et~al.
\newblock 2010.
\newblock A survey on transfer learning.
\newblock {\em IEEE Transactions on knowledge and data engineering}
  22(10):1345--1359.

\bibitem[\protect\citeauthoryear{Pei \bgroup et al\mbox.\egroup
  }{2018}]{pei2018multi}
Pei, Z.; Cao, Z.; Long, M.; and Wang, J.
\newblock 2018.
\newblock Multi-adversarial domain adaptation.
\newblock In {\em AAAI Conference on Artificial Intelligence}.

\bibitem[\protect\citeauthoryear{Perronnin and
  Dance}{2007}]{perronnin2007fisher}
Perronnin, F., and Dance, C.
\newblock 2007.
\newblock Fisher kernels on visual vocabularies for image categorization.
\newblock In {\em 2007 IEEE conference on computer vision and pattern
  recognition},  1--8.
\newblock IEEE.

\bibitem[\protect\citeauthoryear{Saenko \bgroup et al\mbox.\egroup
  }{2010}]{saenko2010adapting}
Saenko, K.; Kulis, B.; Fritz, M.; and Darrell, T.
\newblock 2010.
\newblock Adapting visual category models to new domains.
\newblock In {\em European conference on computer vision},  213--226.
\newblock Springer.

\bibitem[\protect\citeauthoryear{Shen \bgroup et al\mbox.\egroup
  }{2018}]{shen2018wasserstein}
Shen, J.; Qu, Y.; Zhang, W.; and Yu, Y.
\newblock 2018.
\newblock Wasserstein distance guided representation learning for domain
  adaptation.
\newblock In {\em AAAI}.

\bibitem[\protect\citeauthoryear{Simonyan and
  Zisserman}{2014}]{simonyan2014very}
Simonyan, K., and Zisserman, A.
\newblock 2014.
\newblock Very deep convolutional networks for large-scale image recognition.
\newblock {\em arXiv preprint arXiv:1409.1556}.

\bibitem[\protect\citeauthoryear{Sivic and Zisserman}{2003}]{sivic2003video}
Sivic, J., and Zisserman, A.
\newblock 2003.
\newblock Video google: A text retrieval approach to object matching in videos.
\newblock In {\em null},  1470.
\newblock IEEE.

\bibitem[\protect\citeauthoryear{Sun and Saenko}{2016}]{sun2016deep}
Sun, B., and Saenko, K.
\newblock 2016.
\newblock Deep coral: Correlation alignment for deep domain adaptation.
\newblock In {\em European Conference on Computer Vision},  443--450.
\newblock Springer.

\bibitem[\protect\citeauthoryear{Tzeng \bgroup et al\mbox.\egroup
  }{2017}]{tzeng2017adversarial}
Tzeng, E.; Hoffman, J.; Saenko, K.; and Darrell, T.
\newblock 2017.
\newblock Adversarial discriminative domain adaptation.
\newblock In {\em Computer Vision and Pattern Recognition (CVPR)}, volume~1,
  ~4.

\bibitem[\protect\citeauthoryear{Venkateswara \bgroup et al\mbox.\egroup
  }{2017}]{venkateswara2017deep}
Venkateswara, H.; Eusebio, J.; Chakraborty, S.; and Panchanathan, S.
\newblock 2017.
\newblock Deep hashing network for unsupervised domain adaptation.
\newblock In {\em Proc. CVPR},  5018--5027.

\bibitem[\protect\citeauthoryear{Yosinski \bgroup et al\mbox.\egroup
  }{2014}]{yosinski2014transferable}
Yosinski, J.; Clune, J.; Bengio, Y.; and Lipson, H.
\newblock 2014.
\newblock How transferable are features in deep neural networks?
\newblock In {\em Advances in neural information processing systems},
  3320--3328.

\bibitem[\protect\citeauthoryear{Yue-Hei~Ng, Yang, and
  Davis}{2015}]{yue2015exploiting}
Yue-Hei~Ng, J.; Yang, F.; and Davis, L.~S.
\newblock 2015.
\newblock Exploiting local features from deep networks for image retrieval.
\newblock In {\em Proceedings of the IEEE conference on computer vision and
  pattern recognition workshops},  53--61.

\bibitem[\protect\citeauthoryear{Zheng \bgroup et al\mbox.\egroup
  }{2018}]{zheng2018unsupervised}
Zheng, N.; Wen, J.; Liu, R.; ; Long, L.; Dai, J.; and Gong, Z.
\newblock 2018.
\newblock Unsupervised representation learning with long-term dynamics for
  skeleton based action recognition.
\newblock In {\em AAAI},  2644--2651.

\bibitem[\protect\citeauthoryear{Zhou \bgroup et al\mbox.\egroup
  }{2014}]{zhou2014hybrid}
Zhou, J.~T.; Pan, S.~J.; Tsang, I.~W.; and Yan, Y.
\newblock 2014.
\newblock Hybrid heterogeneous transfer learning through deep learning.
\newblock In {\em AAAI},  2213--2220.

\bibitem[\protect\citeauthoryear{Zhou \bgroup et al\mbox.\egroup
  }{2018}]{zhou2018transfer}
Zhou, J.~T.; Zhao, H.; Peng, X.; Fang, M.; Qin, Z.; and Goh, R. S.~M.
\newblock 2018.
\newblock Transfer hashing: From shallow to deep.
\newblock {\em IEEE Transactions on Neural Networks and Learning Systems}.

\end{thebibliography}

\end{document}